\documentclass[12pt]{article}
\usepackage[preprint]{jmlr2e}

\usepackage{amsmath}
\usepackage{amssymb}
\usepackage{amsfonts}

\RequirePackage{natbib}
\RequirePackage{graphicx}

\title{Deep Set Neural Networks for forecasting asynchronous bioprocess timeseries}

\author{\name Maxim Borisyak \email maxim.borisyak@tu-berlin.de\\
	\name Stefan Born \\
	\name Peter Neubauer \\
	\name Mariano Nicolas Cruz-Bournazou \\
	\addr Technische Universität Berlin, Straße 17 des Juni 135, 10623
Berlin, Germany
}

\begin{document}
\maketitle

\begin{abstract}
	Cultivation experiments often produce sparse and irregular time series.
	Classical approaches based on mechanistic models, like Maximum
	Likelihood fitting or Monte-Carlo Markov chain sampling, can easily
	account for sparsity and time-grid irregularities, but most statistical
	and Machine Learning tools are not designed for handling sparse data
	out-of-the-box. Among popular approaches there are various schemes for
	filling missing values (imputation) and interpolation into a regular
	grid (alignment). However, such methods transfer the biases of the
	interpolation or imputation models to the target model.
	
	Following the works of~\citet{yalavarthi2022dcsf}, we show that Deep
	Set Neural Networks~\citep{zaheer2017deep} equipped with triplet encoding
	of the input data can successfully handle bio-process data without any
	need for imputation or alignment procedures. The method is agnostic to
	the particular nature of the time series and can be adapted for any
	task, for example, online monitoring, predictive control, design of
	experiments, etc. In this work, we focus on forecasting.
	
	We argue that such an approach is especially suitable for typical
	cultivation processes, demonstrate the performance of the method on
	several forecasting tasks using data generated from macrokinetic growth
	models under realistic conditions, and compare the method to a
	conventional fitting procedure and methods based on imputation and
	alignment.
\end{abstract}

\section{Introduction}

Experiments in bioprocessing, like in many other areas, often rely on
multiple measuring devices and procedures to probe different quantities
which, most likely, operate with different frequencies. For example, in
bioreactors in the High‐throughput bioprocess development facility used
by~\citet{haby2019integrated}, dissolved oxygen tension and pH are measured much
more frequently than OD\textsubscript{600}, acetate and glucose
concentrations. Measurement times usually are not aligned, which
translates into sparse and irregular timeseries. In the extreme, but not rare cases, only one channel is measured at a time. Conventionally, such data
is analysed with the help of macrokinetic models by either Maximum Likelihood fitting (ML) or Monte-Carlo Markov chain (MCMC) sampling.
In both cases, one utilises the likelihood function which can be easily
formulated for sparse and irregular data.

Consider a bioprocess with two monitored quantities, $A$ and $B$, being monitored, and let
$A\left( t \middle| \theta \right)$ and $B\left( t \middle| \theta \right)$ be of a corresponding macrokinetic
model with parameters $\theta$. Many relevant loss functions that
measure the discrepancy between the data and a model are decomposable
over measurements and therefore are defined for sparse data, e.g., the
negative log likelihood of the data with respect to the macrokinetic
model with additive homoscedastic Gaussian noise:
\begin{align}
	L(\theta) = - \log P(X \mid  \theta) \sim \sum_{i = 1}^{N_A} \left( A_i - A(t^A_i \mid \theta) \right)^2 + \sum_{i = 1}^{N_B} \left( B_i - B(t^B_i \mid \theta) \right)^2;
	\label{eq:loss}
\end{align}
where:
$X = \left( \{ A_{i} \mid i = 1,\ldots, N_{A}\}, \{ B_{j} \mid j = 1, \dots, N_{B}\} \right)$
are measurements of $A$ and $B$ at times $t_{i}^{A}$ and $t_{j}^{B}$ correspondingly.
Thus, both ML and MCMC can naturally handle sparsity and irregularity of the measurement data.

In recent years, Machine Learning algorithms gained popularity in many
areas of natural science, including biotechnology, primarily because of
their ability to learn complex dependencies from data. A wide range of
methods is available in the literature for solving various tasks such as
classification, regression, inference, generation etc. For a general
introduction to Machine Learning we refer readers to \citet{hastie2009elements},
for a review of Machine Learning applications in bioprocess engineering: \citet{duong2023bioprocess, helleckes2023machine, mondal2022review}.

However, the vast majority of supervised Machine Learning algorithms are
designed for complete and homogeneous input data, whereas decomposable
loss functions make them naturally applicable to the prediction of
sparse targets as in Equation~\eqref{eq:loss}. From a Machine Learning point of view
fitting the parameters and initial conditions of a macrokinetic growth
model is a regression from time points to observations. A fitted models
can then be used to forecast observations, but forecasting can also be
framed as a genuine machine learning problem with past observations as
inputs and future observations as targets, more precisely: given a set
of sparse and irregular observations
$X = \left\{ \left( t_{i},x_{i} \right) \mid i = 1, \dots, N \right\}$
generated by a process with unknown parameters (e.g., unknown strain of
bacteria), predict values in the timepoints
$T = \left\{ \tau_{j} \mid j = 1, \dots, M \right\}$. Other
examples of machine learning problems with a time series of
observations as input are inference: predicting macrokinetic
parameters given observations; optimal control: predicting controls to
steer cultivations towards a predetermined target. Algorithms for these
types of tasks do not accept missing values ``out-of-the-box''. Methods
based on fitting and MCMC effectively ``train'' a new regressor for each
set of measurements, then use this regressor for making predictions or
optimising controls. At the same time, one can rarely afford to train a
Machine Learning model for each cultivation since a typical amount of
measurements is not sufficient to properly train even a small
general-purpose model. Instead, the typical approach is to train a model
on data from a large number of different processes (varying initial
conditions, different strains or species, different experimental
parameters, etc.), effectively learning a direct mapping from
measurements to predictions.

\begin{figure}[t]
	\centering
	\includegraphics[width=\textwidth]{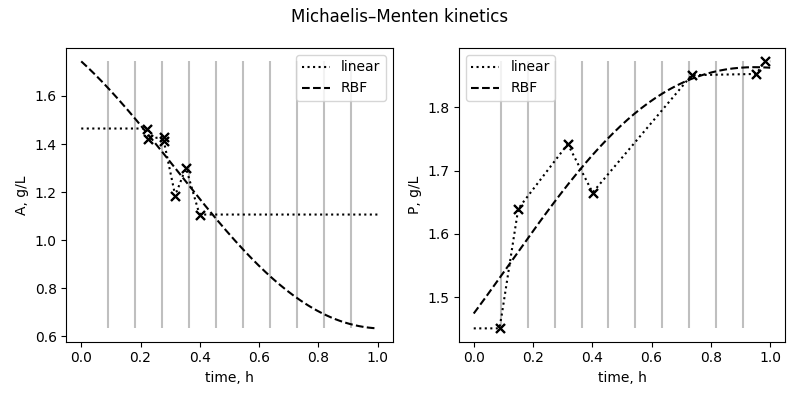}
	\caption{Interpolation methods on irregular and sparse data sampled from Michaelis-Menten kinetics model: “linear” – linear spline, “RBF” – RBF kernel regression; A denotes substrate, P – product. Alignment and imputations are done by projecting predictions into a regular grid (grey vertical lines), the resulting values are omitted to reduce visual clutter.}
	\label{fig:imputation}
\end{figure}

A large number of heuristical procedures were developed for dealing with
sparse inputs. Most of them focus on filling the missing measurements
with some concrete values, i.e., imputation \citet{carpenter2023multiple},
for example, by replacing missing values with average value for the
channel. For timeseries, imputation can be seen as a regression problem,
i.e., filling missing values with predictions of a regressor trained on
the available data. Aligning data to a regular grid can be considered as
a special case of imputation. In this work we consider two methods that work well with
irregular timeseries: fitting splines or kernel regression to each
channel of each sample and replacing missing measurements with
predictions of the models on a predetermined regular time grid (Figure~\ref{fig:imputation}).
The major downside of this approach is that biases and errors introduced by imputation and alignment are translated into the model
trained on the imputed data. In this work we show that for Deep Learning
models imputation and alignment are unnecessary steps.

\section{Deep Learning for Sparse Data}

Although the majority of Machine Learning methods are formulated for
complete data, there is a growing body of literature on methods for
sparse data, especially for processing timeseries. Some of the methods
are based on variations of Neural ODE~\citep{chen2018neural}: the model is
defined by an ODE where RHS is represented by a neural network~\citet{rubanova2019latent}.
Others mimic continuous dynamics via Recurrent Neural
Networks with updates dependent on the time step~\citep{de2019gru, schirmer2022modeling}. In both cases, the continuous nature of
predictions allows handling irregular samples in a natural manner. In
this work we focus on another popular approach, namely, Deep Set
Networks~\citet{zaheer2017deep}.

Deep Set Networks were designed to model set functions, i.e., functions
invariant to input permutations. They achieve the invariance by
introducing two subnetworks: extractor
$f:{\mathbb{R}}^{n} \rightarrow {\mathbb{R}}^{m}$ and aggregator
$g:{\mathbb{R}}^{m} \rightarrow {\mathbb{R}}^{l}$. For a set of
observations
$X = \left\{ x_{i} \in {\mathbb{R}}^{n} \mid i = 1, \dots, N \right\}$
the network's output $h\left( X \right)$ is defined as follows:
\begin{align}
h\left( X \right) = g\left( \sum_{i = 1}^{N}f\left( x_{i} \right) \right). \label{eq:deepset}
\end{align}
The extractor network translates each of the samples into a latent
representation which is then summed with others. The aggregator then
translates the sum into predictions. As can be seen from Equation~\eqref{eq:deepset}, the
output of the network does not depend on the particular order of the
observations as the order is ``erased'' by the summation. Of course, the
order of observations is hugely important in forecasting. Equipping
observations with timestamps:
\begin{align*}
	X = \left\{ \left( t_{i},x_{i} \right) \mid i = 1,\text{...},N \right\},
\end{align*}
makes the observations invariant to permutations of the index
while preserving the time structure.

\begin{figure}[t]
	\centering
	\includegraphics[width=\textwidth]{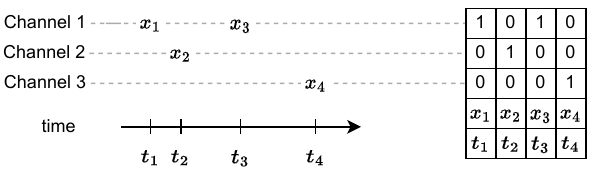}
	\caption{Interpolation methods on irregular and sparse data sampled from Michaelis-Menten kinetics model: ``linear'' --- linear spline, ``RBF'' --- RBF kernel regression; $A$ denotes substrate, $P$ --- product. Alignment and imputations are done by projecting predictions into a regular grid (grey vertical lines), the resulting values are omitted to reduce visual clutter.}
	\label{fig:triplet}
\end{figure}

To employ Deep Sets for forecasting, we adopt a structure similar to
that of Conditional Neural Processes~\citep{garnelo2018conditional}: the
aggregator is additionally supplied with a timestamp $\tau$ in which
predictions $p$ are requested:
\begin{align*}
p = g\left( \tau,\sum_{i = 1}^{N}f\left( x_{i},t_{i} \right) \right).
\end{align*}

For example, for a cell cultivation timeseries, the extractor network,
most likely, captures important macrokinetic parameters of cell growth,
albeit, in a not readily interpretable form. It is worth noting that not
all statistics can be captured with summation~\citep{wagstaff2019limitations},
however, with an upper limit on the number of observations and
sufficiently large dimensionality of the latent space, one could
approximate any permutation invariant statistics with an arbitrary
precision.

We represent timeseries by the so-called triplet format~\citep{horn2020set, yalavarthi2022dcsf}, which encodes each single measurement
by the tuple of timestamp, channel indicator and measured value. It
makes no difference whether one or several channels are measured at a
specific time point, in particular there is no need for special
treatment of missing values. The process is illustrated in Figure~\ref{fig:triplet}.

\section{Numerical experiments}

To evaluate the performance of the proposed method we simulate data
from two mechanistic models: Michaelis--Menten kinetics (MMK) and E.
coli growth model by \citet{anane2017modelling}, both expressed as ODE
systems. For each model we sample parameters from a wide prior
distribution, obtain trajectories and from each trajectory we generate
a small number of observations for each of the observed channels. We
emulate a scenario in which networks are presented with observations
until a certain horizon (``training'' part), and asked to predict
future states of the system (``forecasting'' part). For MMK we sample
14 observations per part, for the E. coli model: 30. To test the most
extreme scenario, all measurement timestamps are randomly distributed
within their corresponding ranges (thus, highly irregular), and the
datasets contain only single-channel measurements.

\begin{figure}[t]
	\centering
	\begin{tabular}{| l | l | l |}
	\hline
	Method & MMK, $R^2$ & E. coli, $R^2$\\
	\hline
	Ground truth & $0.9797 \pm 6\mathrm{E}{-5}$ & $0.994 \pm 1\mathrm{E}{-5}$\\
	\hline
	Mechanistic Model & $0.9073 \pm 0.0020$ & $0.387 \pm 0.03$\\
	\hline
	Deep Sets + linear splines & $0.9431 \pm 0.0004$ & $0.848 \pm 0.006$ \\
	\hline
	Deep Sets + RBF kernel reg. & $0.9446 \pm 0.0004$ & $0.835 \pm 0.007$ \\
	\hline
	Deep Sets + triplet encoding & $0.9481 \pm 0.0003$ & $0.856 \pm 0.006$ \\
	\hline
	\end{tabular}
	\caption{
		Results of the experiments. ``Mechanistic Model'' is fitted to the data, ``ground truth'' denotes the corresponding mechanistic model evaluated with the true parameters. Squared errors are normalized by noise variances for the corresponding channel.
	}
	\label{fig:results}
\end{figure}

We compare Deep Sets with triplet encoding against neural networks
equipped with imputation and alignment procedures based on linear
splines and kernel regression. To produce networks of a similar size, we
also employ Deep Sets architecture for the imputation-based networks;
they, however, receive fully imputed observations on a regular grid of
the same size as the number of observations. As a baseline, we use the
corresponding mechanistic models fitted to the measurements with BFGS
optimisation algorithm~\citet{fletcher2000practical}. For comparison, we also
evaluate mechanistic model using true parameters, results of which
indicate loss due to measurement noise.

The results are summarised in Figure~\ref{fig:triplet}. Surprisingly, all Machine
Learning models surpass fitting with mechanistic models, especially for
the E. coli model. We believe the mean reason behind this poor
performance is the local minima problem. Machine Learning algorithms
seem unaffected by such a problem. Deep Set Networks with triplet
encoding are either statistically better or on par with the methods
based on imputation and alignment, however, the difference is
practically negligible.

\section{Conclusion}

Using Deep Set Networks as an example, we demonstrate experimentally
that imputation and alignment procedures are unnecessary for Deep
Learning models when dealing with sparse and irregular data.
Additionally, we showed that such models surpass conventional fitting
procedures, even when the data is generated from the same underlying
macrokinetic model.

\section*{Acknowledgements}

We gratefully acknowledge the financial support of the German Federal Ministry of Education and Research (01DD20002A -- KIWI biolab).

\bibliography{DeepSetsForecasting}

\end{document}